\pdfoutput=1
\documentclass[letterpaper, 10 pt, conference]{ieeeconf}  % Comment this line out if you need a4paper

\IEEEoverridecommandlockouts                              % This command is only needed if 
\usepackage{epsfig} % for postscript graphics files
\usepackage{graphicx}
\usepackage{pifont}
\usepackage{url}
\usepackage{here}
\usepackage{cite}
\usepackage{amsmath} % assumes amsmath package installed
\usepackage{amssymb}  % assumes amsmath package installed
\usepackage{multirow}

\title{\LARGE \bf
Unsupervised Simultaneous Learning for Camera Re-Localization \\ and Depth Estimation from Video
}

\author{Shun Taguchi$^{1}$ and Noriaki Hirose$^{1}$% <-this % stops a space
\thanks{*This work was not supported by any organization}% <-this % stops a space
\thanks{$^{1}$Shun Taguchi and Noriaki Hirose are with Toyota Central R\&D Labs., Inc., 41-1 Yokomichi, Nagakute, Aichi, Japan
        {\tt\small s-taguchi@mosk.tytlabs.co.jp}
        {\tt\small hirose@mosk.tytlabs.co.jp}
        }%
}

\begin{document}

\maketitle
\thispagestyle{empty}
\pagestyle{empty}

%%%%%%%%%%%%%%%%%%%%%%%%%%%%%%%%%%%%%%%%%%%%%%%%%%%%%%%%%%%%%%%%%%%%%%%%%%%%%%%%
\begin{abstract}
We present an unsupervised simultaneous learning framework for the task of monocular camera re-localization and depth estimation from unlabeled video sequences.
Monocular camera re-localization refers to the task of estimating the absolute camera pose from an instance image in a known environment, which has been intensively studied for alternative localization in GPS-denied environments.
In recent works, camera re-localization methods are trained via supervised learning from pairs of camera images and camera poses.
In contrast to previous works, we propose a completely unsupervised learning framework for camera re-localization and depth estimation, requiring only monocular video sequences for training.
In our framework, we train two networks that estimate the scene coordinates using directions and the depth map from each image which are then combined to estimate the camera pose.
The networks can be trained through the minimization of loss functions based on our loop closed view synthesis.
In experiments with the 7-scenes dataset, the proposed method outperformed the re-localization of the state-of-the-art visual SLAM, ORB-SLAM3.
Our method also outperforms state-of-the-art monocular depth estimation in a trained environment.
\end{abstract}

%%%%%%%%% BODY TEXT
\section{Introduction}
\begin{figure}
\begin{center}
\includegraphics[width=1.0\hsize]{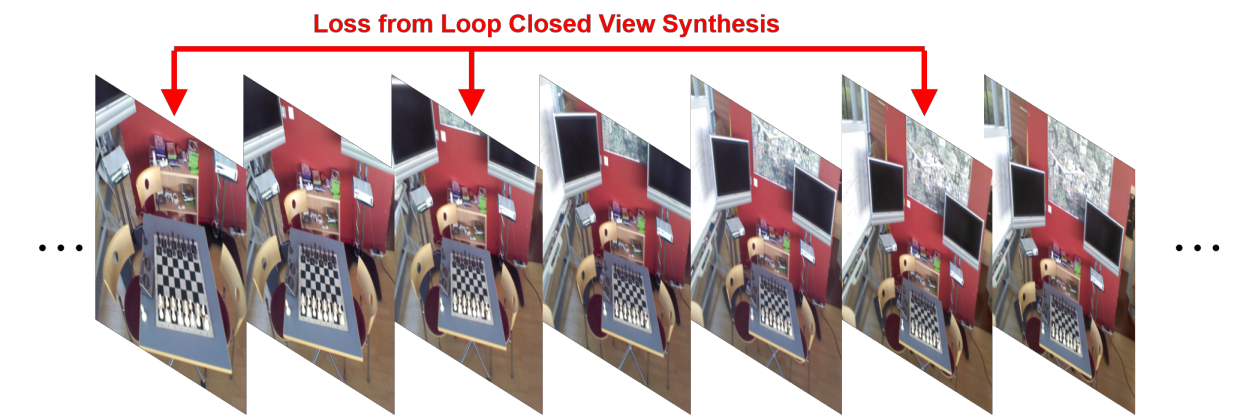}
{\small (a) Training from unlabeled monocular video sequences.}
\includegraphics[width=1.0\hsize]{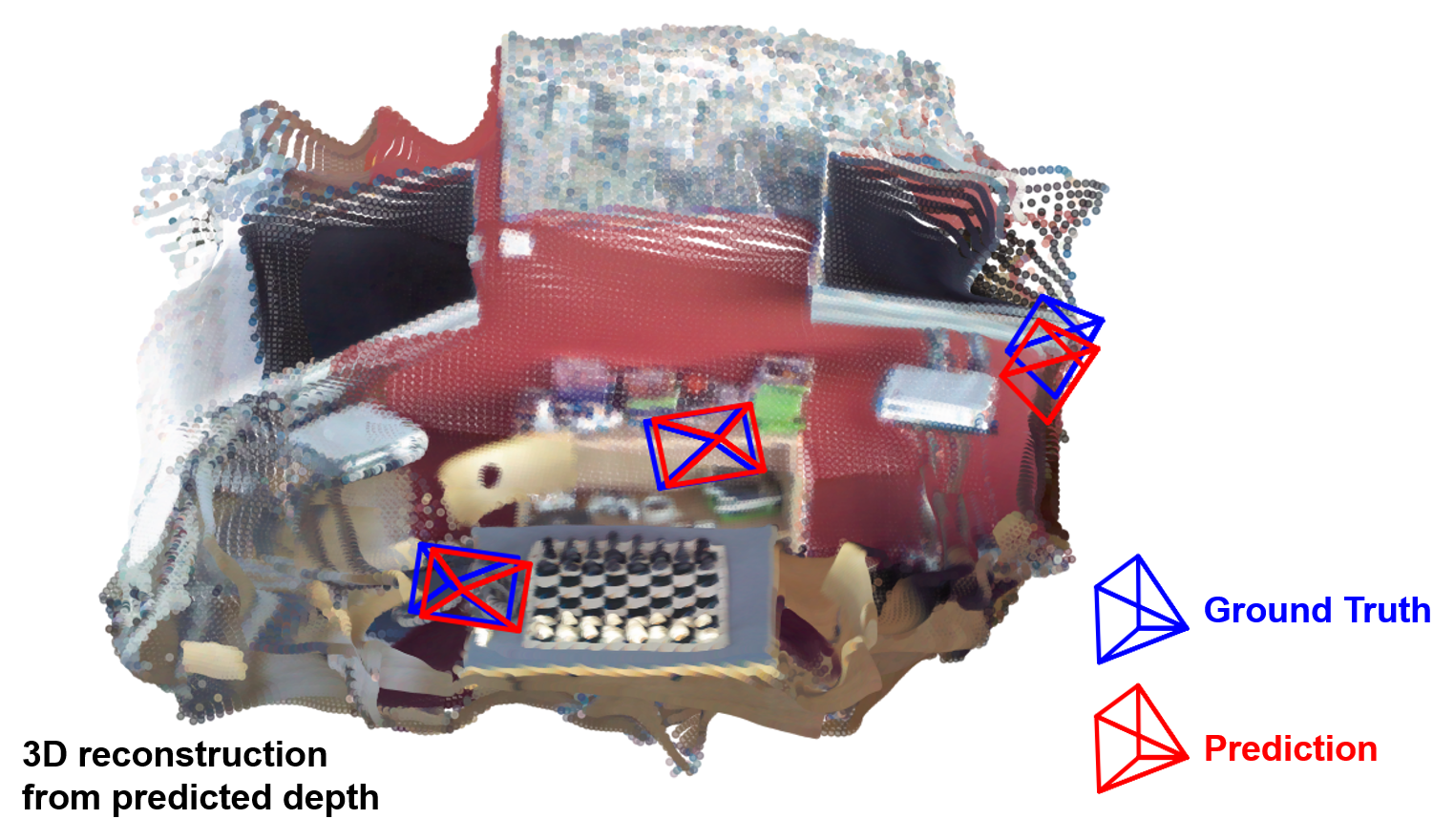}
{\small (b) Test: camera re-localization (camera pose estimation from a single monocular camera image) with depth estimation.}
\caption{{\bf Unsupervised simultaneous learning for camera re-localization and depth estimation.}
We present an unsupervised simultaneous learning framework for camera re-localization and depth estimation from unlabeled monocular video sequences.
Our models predict the camera pose and depth in a known environment at test time.
}
\label{fig:pull}
\end{center}
%\vspace{-4mm}
\end{figure} 

Relying on a single memory image, humans can easily recognize a previously visited environment.
Humans can also comprehend the implicit map of any environment relying on only the memory of their first visit. Immediately they see a new image, they can recognize where it belongs on the map.
How can humans construct an implicit map in their brains relying only on unlabeled image sequences? 
We consider that humans recognize relative pose and depth from the difference of appearance in the images. They also perform re-localization on the images to recognize a place as the same place when they see it for a second time.
Inspired by this human ability, we propose a novel unsupervised learning framework to train neural networks to estimate camera pose from a single image.

To this end, we trained camera re-localization and depth estimation simultaneously from unlabeled monocular video sequences by utilizing view synthesis inspired by monocular depth estimation~\cite{zhou2017unsupervised,monodepth2}.
By introducing re-localization networks, it is possible to construct implicit maps while determining that places with the same appearance are the same places.
Our training led the networks to implicitly understand the environment captured in the training dataset, thereby, achieving accurate camera re-localization and depth estimation in inference (see Fig. \ref{fig:pull}).

Our method can train networks for localization and structure understanding of the environment via a single learning process.
This information is essential for visual-based autonomous navigation of robots.
Our method is smart and useful because it can simultaneously learn the networks required for such autonomous navigation.

This study makes three main contributions:
\begin{itemize}
      \item We develop a novel unsupervised simultaneous learning framework for camera re-localization and depth estimation from unlabeled video sequences.
      \item We propose a novel directed scene coordinate network that enables accurate estimation of the camera pose from a single image by combining it with an estimated depth map.
      \item We propose a calculation process of the loss function with loop closed view synthesis, which improves the training of camera re-localization.
\end{itemize}
To the best of our knowledge, our study is among the first studies to propose unsupervised camera re-localization by training models from unlabeled video sequences and estimating camera pose from a single image in inference.

The proposed method outperformed the re-localization systems embedded in the state-of-the-art visual SLAM method, ORB-SLAM3 \cite{orbslam3}.
Our method also outperformed one of the unsupervised monocular depth estimation methods, monodepth2~\cite{monodepth2} in depth estimation. Our method can suppress the deviation of scale in the test dataset on depth estimation because more consistent scales are learned via camera pose estimation. 
These evaluations are quantitatively and qualitatively conducted on 7-scenes dataset~\cite{shotton2013scene}.

\section{Related Work}

In the following, we discuss camera re-localization and monocular depth estimation, which are most closely related to our study.

\subsection{Camera Re-Localization}

In early studies of camera re-localization based on image retrieval \cite{schindler2007city}, images were retrieved from a database by matching query images with global image descriptors \cite{torii201524, arandjelovic2016netvlad}.
Because the re-localization accuracy of these retrieval approaches is limited by the sampling density of database images, various camera re-localization methods have been proposed to address this problem.

In addition, absolute pose regression methods \cite{walch2017, posenet, kendall2017geometric, naseer2017deep, brahmbhatt2018geometry, laskar2017camera, valada2018deep} have been used to train neural networks to regress the camera pose using database images as the training set.
However, in practice, absolute pose regression methods do not outperform the accuracy of image retrieval methods \cite{sattler2019understanding}.
Relative pose regression methods \cite{relocnet, sattler2019understanding} train a neural network to predict the relative transformation between the query image and the database image most similar to that which is obtained by image retrieval.
A recent study \cite{camnet} suggested that relative pose regression can achieve accuracy comparable to that of the structure-based methods described below.

Structure-based approaches, such as scene coordinate regression \cite{shotton2013scene}, are some of the most successful approaches.
Scene coordinate regression \cite{shotton2013scene} directly predicts the 3D scene points corresponding to a given 2D pixel location, and the camera pose is calculated by solving the PnP problem.
Originally, scene coordinate regression was proposed for RGB-D-based re-localization in indoor environments \cite{shotton2013scene, valentin2015exploiting, guzman2014multi, meng2018exploiting}.
Recently, scene coordinate regression has been shown to be effective for RGB-based re-localization \cite{brachmann2016uncertainty,meng2017backtracking}.
DSAC++ \cite{brachmann2018learning} presents the possibility of learning scene coordinate regression from RGB images and ground-truth poses only.
In subsequent work, DSAC* \cite{dsacstar} makes several improvements to DSAC++ to achieve state-of-the-art performance.

Recently, some studies have extended the camera re-localization method to the time domain in order to address temporal re-localization \cite{clark2017vidloc, valada2018deep, radwan2018vlocnet++, xue2019local, zhou2020kfnet}.  
While these approaches are effective for some applications, we will focus on one-shot camera re-localization because it is a more fundamental and widely used technique.

The reviewed literature referred to above has focused on supervised learning of camera re-localization.
In contrast, we present an unsupervised framework for camera re-localization using unlabeled video sequences.
We also estimate the scene coordinates based on the findings of these studies of camera re-localization.
Unlike \cite{brachmann2018learning,dsacstar}, when solving the PnP problem, our method estimates the scene coordinate, which is a 6-dimensional coordinate with the gaze direction to handle the difference in appearance depending on the viewing direction.
Our approach to estimating the camera pose is introduced in our unsupervised learning framework based on loop closed view synthesis, which is inspired by unsupervised monocular depth estimation.

\subsection{Monocular Depth Estimation}

One of the earliest works in convolutional-based depth estimation was presented by Eigen et al \cite{eigen2014depth}. Eigen et al used a multi-scale deep network trained on RGB-D sensor data to regress the depth directly from single images.
Inspired by the two-view stereo disparity estimation \cite{mayer2016large} based on flow estimation \cite{dosovitskiy2015flownet}, Umenhofer et al. \cite{ummenhofer2017demon} trained a depth and pose network simultaneously to predict depth and camera ego-motion between successive unconstrained image pairs. To address the difficulty of labeling the target depth,  \cite{garg2016unsupervised, godard2017unsupervised} trained a monocular depth network with stereo cameras without requiring ground-truth depth labels.
Godard et al. \cite{godard2017unsupervised} used stereo images to geometrically transform the right image into a left image based on a predicted depth by leveraging spatial transformer networks \cite{jaderberg2015spatial}.
The photometric re-projection loss between the synthesized and original left images can be used to train the depth network without the ground-truth depth.

Following \cite{godard2017unsupervised} and \cite{ummenhofer2017demon}, Zhou et al. \cite{zhou2017unsupervised} applied this unsupervised training to a purely monocular setting, where a depth and relative pose network are simultaneously learned from unlabeled monocular videos.
Recent studies \cite{casser2019depth, klodt2018supervising, mahjourian2018unsupervised, wang2018learning, zhou2018unsupervised, zou2018df,hirose2021plg,hirose2021variational} have incorporated these methods, additional loss, and constraints.
Monodepth2 \cite{monodepth2} is a successful method that employs a ResNet encoder \cite{he2016deep} and has achieved state-of-the-art performance.
More recent methods employ improved network models with more parameters to achieve state-of-the-art performance \cite{mallya2018packnet}. 
Several methods \cite{yang2018deep, yin2018geonet, liang2021deep} have focused on pose estimation based on an unsupervised monocular depth estimation framework, however, these approaches have focused only on relative pose estimation between two images, that is, visual odometry.

Instead of the visual odometry network of unsupervised monocular depth estimation, our method estimates the absolute camera pose in an implicit map via estimated scene coordinates, which is trained with our loop closed view synthesis. 
Note that the estimated camera pose in our method is not a relative pose between two images.
To the best of our knowledge, our method is the first of its kind to train camera re-localization in an unsupervised setting.

\section{Method}
\begin{figure*}
\begin{center}
\includegraphics[width=1.0\hsize]{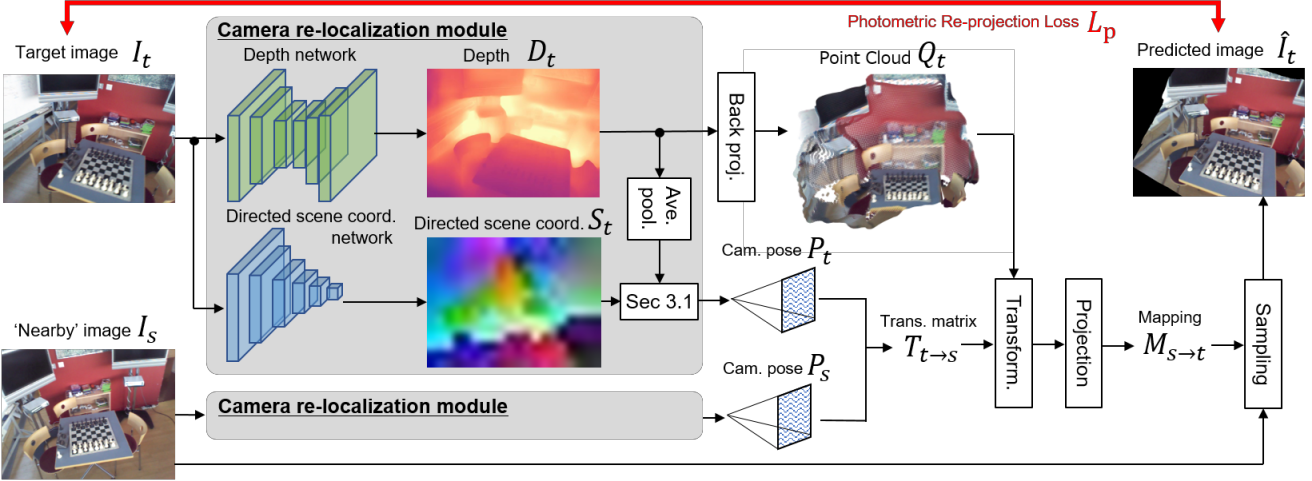}
\caption{{\bf Overview of our unsupervised simultaneous learning framework of directed scene coordinate and depth networks.}
Our model consists of two networks for depth and directed scene coordinate.
The camera pose can be calculated geometrically from a directed scene coordinate and depth as shown in Sec.\ref{seq:camera_reloc}, and the relative pose between an image $I_t$ and a `nearby' image $I_s$ can be calculated from the predicted camera pose from each image independently.
By using the relative pose and back-projected point cloud from depth, the corresponding map $M_{s \rightarrow t}$ can be calculated.
Our networks are trained by the photometric re-projection loss $L_p$ between the original $I_t$ and the view synthesis image $\hat{I}_t$ from the `nearby' image $I_s$.
}
\label{fig:overview}
\end{center}
\vspace{-5mm}
\end{figure*}

\subsection{\label{seq:camera_reloc}Camera Re-Localization based on Directed Scene Coordinate and Depth}

\begin{figure}
\begin{center}
\includegraphics[width=1.0\hsize]{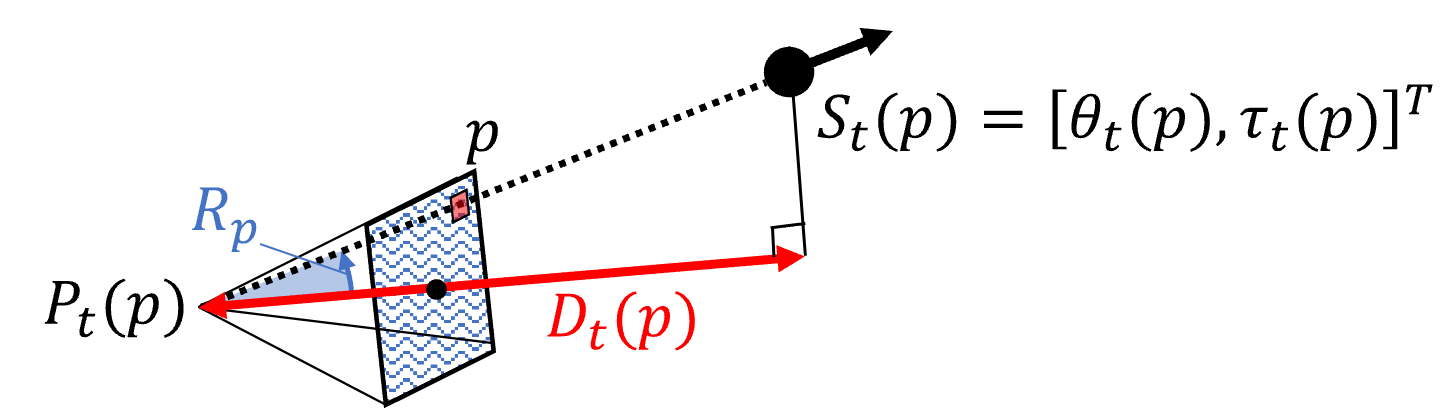}
\caption{{\bf Camera pose calculation from directed scene coordinate and depth.}
The camera pose $P_t (p)$ can be calculated from a pixel of the directed scene coordinate $S_t (p)$ and the corresponding depth $D_t (p)$ by rotating the pose based on the direction of the pixel and translating the position based on the depth and direction of the scene coordinates.
The final pose is obtained as the average of estimated camera poses from each pixel.
}
\label{fig:camera_pose_calculation}
\end{center}
\vspace{-5mm}
\end{figure}

In this study, we propose a novel unsupervised simultaneous learning framework for camera re-localization and depth estimation, shown in Fig. \ref{fig:overview}.
Our model consists of two networks for depth and directed scene coordinate.
The depth network predicts the dense depth map $D_t$ from a single RGB image $I_t$ at time step $t$.
The directed scene coordinate network predicts the direct scene coordinate $S_t$, which consists of subsampled scene coordinates with direction from a single RGB image $I_t$.
The directed scene coordinate $S_t = [\theta_t, \tau_t]^T$ consists of a 3-dimensional attitude $\theta_t$ as an axis-angle representation and 3-dimensional position $\tau_{t}$.
Here, $\theta_t$ indicates the direction of the vector from the camera position to the position of corresponding scene coordinates.

The camera pose $P_t$ can be calculated geometrically from a directed scene coordinate $S_t$ and the depth $D_t$ for each pixel.
To calculate camera pose, we introduce the point cloud $Q_t$, which can be obtained by back projection from the depth $D_t$ as follows:
\begin{align}
Q_t (p) &= D_t (p) K_t^{-1} \left[ \begin{array}{c}
p \\
1
\end{array} \right],
\end{align}
where $Q_t (p)$ is a vector to local point corresponding to pixel $p = [u, v]^T$, and $K_t$ is the camera intrinsic parameter for $I_t$.

The camera pose $P_t (p)$ corresponding to a pixel $p$ also consists a 3-dimensional attitude $\theta_{P_t} (p)$ as an axis-angle representation and a 3-dimensional position $\tau_{P_t} (p)$ as
\begin{align}
P_t (p) &= \left[ \begin{array}{c}
\theta_{P_t} (p) \\
\tau_{P_t} (p) 
\end{array} \right], 
\end{align}
and it is calculated from the directed scene coordinate $S_t (p)$ and the point $Q_t (p)$ (see. Fig. \ref{fig:camera_pose_calculation}).
The camera attitude $\theta_{P_t} (p)$ is calculated by rotating the gaze direction $\theta_t (p)$ using the direction of the pixel $p$ corresponding to the directed scene coordinates. 
\begin{align}
R(\theta_{P_t}(p)) &= R_{p}^{-1} R(\theta_t (p))
\end{align}
where $R_{p}$ is the rotation matrix of gaze direction to the pixel $p$, 
$R(\cdot)$ represents rotation matrix corresponding the 3-dimensional attitude.
The camera position $\tau_{P_t} (p)$ is calculated by translating the scene coordinates $\tau_t (p)$ by the distance of the local point $Q_t (p)$ in the gaze direction $\theta_t (p)$.
\begin{align}
\tau_{P_t}(p) &= \| Q_t (p) \| \cdot R(\theta_t (p)) e_z + \tau_t (p),
\end{align}
where $e_z$ is a unit vector to depth direction, and $\|Q_t (p)\|$ is the distance of the point $Q_t (p)$

In this procedure, the camera poses are calculated from all pixels of the directed scene coordinate; therefore, the final pose $P_t$ is obtained as the average (median in testing) of the estimated camera poses from all pixels.
\begin{align}
\label{eq:camera_pose}
P_t = \frac{1}{n_S} \sum_{p} P_t (p),
\end{align}
where, $n_S$ is the number of pixels of directed scene coordinate.

\subsection{\label{sec:loss}Unsupervised Learning based on Photometric Re-projection Loss}

Here, we propose a framework for jointly training two networks that estimate directed scene coordinates and depths from unlabeled video sequences.
Each model can be used independently during test-time inference, and camera re-localization can be performed by combining the two networks, as shown in Sec. \ref{seq:camera_reloc}.

Our models were trained from image sequences obtained from moving cameras in the target environment.
We assume that the corresponding environments are mostly static, that is, the scene appearance change across different frames is dominated by the camera motion.
Similar to most unsupervised monocular depth estimations \cite{zhou2017unsupervised, monodepth2}, we also formulate our problem as the minimization of a photometric re-projection error at training time.
Geometrically, a pixel mapping $M_{s \rightarrow t}$ between $I_t$ and $I_s$ can be derived from the depth $D_{t}$ and $T_{t \rightarrow s}$ as follows~\cite{zhou2017unsupervised, monodepth2}:
\begin{align}
M_{s \rightarrow t} = K_s T_{t \rightarrow s} D_{t} K_t^{-1},
\end{align}
where $K_t$ and $K_s$ are the camera intrinsic matrices for $I_t$ and $I_s$, respectively.
The transformation matrix $T_{t \rightarrow s}$ between two images can be calculated from estimated camera poses $P_t$ and $P_s$ from each image independently as
\begin{align}
\label{eq:relative_pose}
T_{t \rightarrow s} = T(P_s)^{-1} T(P_t),
\end{align}
where $T(\cdot)$ denotes a transformation matrix corresponding to a 6-DoF pose.

By using the relative pose between two images, the base of our unsupervised learning framework for camera re-localization can be formulated.
The synthesized image $\hat{I}_t$ can be constructed by sampling from $I_s$ by following $M_{s \rightarrow t}$.

Similar to previous works \cite{monodepth2}, we employ L1 and SSIM loss as a photometric re-projection loss, which is formulated as
\begin{align}
L_p =& \frac{1}{|V|} \sum_{p \in V}{
	\alpha \frac{
		(1 - SSIM(I_{t}(p), \hat{I}_{t}(p)))
	}{2}
} \nonumber\\
& + \frac{1}{V} \sum_{p \in V}{
	(1 - \alpha) \| I_{t}(p) - \hat{I}_{t}(p) \|_1
},
\end{align}
where $V$ is a set of valid points $p$ that are successfully transformed between $I_t$ and $I_s$,
and $|V|$ denotes the number of points in $V$.
The parameter $\alpha = 0.85$ to follow previous works \cite{monodepth2}.

The photometric re-projection loss is effective for texture-rich scenes, however fragile for low-texture or homogeneous regions.
Therefore, another smoothness loss $L_s$ is used together with the photometric re-projection losses to solve this problem \cite{monodepth2},
\begin{align}
L_s = \frac{1}{n_D} \sum_{p}{\left(e^{-\nabla I_t(p)} \cdot \nabla D_t(p) \right)^2},
\end{align}
where $n_D$ is the number of pixels of $D_t$, and $\nabla$ denotes the first derivative.
This ensures that the smoothness of the depth map is constrained by the primary gradient of the corresponding color image. Following \cite{monodepth2}, $L_p$ and $L_s$ are calculated for each estimated multi-scale depth at the resolution of each decoder's layer.

In this study, we adopt an additional loss to these well-known losses in unsupervised depth estimation.
The loss is the pose coordinate loss $L_c$, which is the error between the camera pose estimated from each pixel and the average final pose in \eqref{eq:camera_pose}.
\begin{align}
L_c = \frac{1}{n_S} \sum_{p}{\|P_t - P_t (p)\|_2}. 
\end{align}

Therefore, the final training loss $L$ consists of photometric re-projection loss, smoothness loss, and pose coordinate loss, as follows:
\begin{align}
L = L_p + w_s L_s + w_c L_c,
\end{align}
where $w_s$ and $w_c$ are the respective weight parameters of smoothness loss and pose coordinate loss. 

\subsection{Loop Closed View Synthesis}
\begin{figure}
\begin{center}
\includegraphics[width=0.7\hsize]{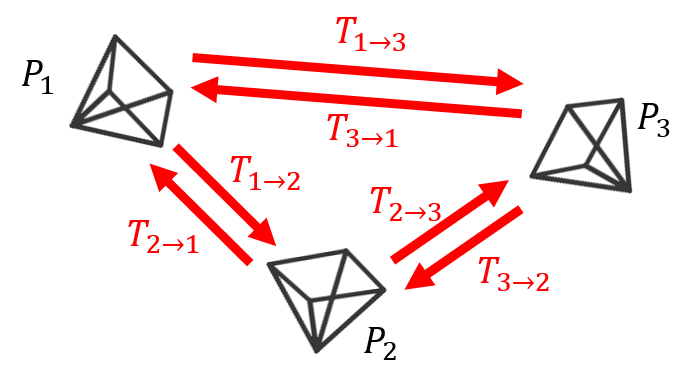}
\caption{{\bf Loop closed view synthesis.}
The monocular depth estimation \cite{monodepth2} learn from view synthesis based on $T_{1 \rightarrow 2}$ and $T_{3 \rightarrow 2}$, however, the two relative poses are insufficient to estimate the three absolute camera poses on an implicit map.
To address this issue, we train our networks from view synthesis for all combinations of images.
}
\label{fig:combination}
\end{center}
%\vspace{-5mm}
\end{figure}
In this study, to achieve effective training for camera pose, we introduced loop closed view synthesis.
First, we picked three or more images from sequences.
For each combination, we bi-directionally predict the images by view synthesis, following Sec.\ref{sec:loss}, and calculate the photometric reprojection loss.
According to our loop closed view synthesis, we calculate six photometric reprojection losses and average them as $L_p$ for three image cases (see Fig.\ref{fig:combination}).

This is because the photometric reprojection loss from two images is insufficient to train networks for camera re-localization.
Eq. \eqref{eq:relative_pose} provides one constraint for two camera poses, which is insufficient to estimate absolute camera poses on an implicit map.
Massive iteration in the learning process may provide sufficient constraints; however, the training process becomes unstable and does not converge smoothly.
Our loop closed view synthesis attempts to solve this issue.
Sufficient constraints to estimate the absolute camera pose stabilize the training process of unsupervised camera re-localization and leads to adequate results.

The final training loss $L$ is averaged over the scale, batch, and view synthesis.

\subsection{Network Architecture}
% Depth Network
Our depth network is the same as in \cite{monodepth2}.
The network is based on the general U-Net architecture \cite{ronneberger2015u}, that is, an encoder-decoder network, with skip connections, enabling us to represent both deep abstract features as well as local information.
We also used ResNet18 \cite{he2016deep} as the depth encoder and started with weights pre-trained ImageNet \cite{russakovsky2015imagenet}.
The depth decoder has sigmoids at the output and we convert the sigmoid output $\sigma$ to depth with $D = 1 / (a \sigma + b)$, where $a$ and $b$ are chosen to constrain $D$ between $0.1$ and $100$ units, following \cite{monodepth2}.

% Directed Scene Coordinate Network
Our directed scene coordinate network also consists of an encoder-decoder network.
We also use a pre-trained ResNet18 as the directed scene coordinate encoder.
The decoder consists of three CNN layers with Rectified Linear Unit (ReLU) activations.
The output of the decoder is the directed scene coordinates $S_t$ with 6 channels of 20$\times$15 matrix. 
The output directed scene coordinate is subsampled by factor 32 from the original images (640$\times$480).
Therefore, we resize the estimated depth by $32 \times 32$ average pooling layer to combine with directed scene coordinates for calculating the camera pose.

\section{Experiments}

\subsection{Dataset}

In this study, an indoor camera dataset, 7-scenes dataset~\cite{shotton2013scene}, was used for the evaluation.
The 7-scenes dataset is an RGB-D indoor re-localization dataset of seven small indoor environments with challenging conditions such as motion blur, reflective surfaces, repetitive structures, and untextured areas.
The dataset was collected using KinectFusion \cite{izadi2011kinectfusion} to record the RGB image, the depth map, and the estimated camera pose.
For each scene, several sequences with 500 or 1000 frames are available, splitting into training and test sets.
We trained our network using only RGB images, essentially, we did not use depths and camera poses for training.
For testing, the provided camera poses are used as the pseudo-ground-truth poses for evaluation. 

\subsection{Training}
At the training time, we select three images to take our loop closed view synthesis.
An image is picked at random from whole train sequences, and the other two `nearby' images are picked within 20 steps before or after the target image.
Each image is scaled in the range of $1.0 \sim 1.1$ and cropped by its original size ($640 \times 480$) as a data augmentation.
We also augment its color via random brightness, contrast, saturation, and hue jitter with respective ranges of $\pm 0.2$, $\pm 0.2$, $\pm 0.2$, and $\pm 0.1$.
Importantly, the color augmentations are only applied to the images that are fed to the networks, and not to those used to compute $L_p$.
The weight parameters of the losses are set as $w_s = 0.001$ and $w_c = 0.03$.

The training is performed through a training split for each scene of the 7-scenes dataset \cite{shotton2013scene}.
The models were trained within 300 epochs with a batch size of 6, and the `nearby' images were selected at random from whole sequences for 50 \% in the latter 100 epochs to learn consistency between images that are distant in time series.
During training, we used batch normalization \cite{ioffe2015batch} for all the layers except for the output layers, and the Adam \cite{kingma2014adam} optimizer with $\beta_1 = 0.9$, $\beta_2 = 0.999$, and a learning rate of $0.0001$.
The training was executed on TITAN RTX GPUs.
We implemented all our systems using the PyTorch \cite{paszke2017automatic} framework.
In our setting, the training process takes about 50 sec per one training image.

\subsection{Results}

\begin{table*}
\caption{{\bf Median re-localization error of position (m) and attitude ($^\circ$) on the 7-scenes test dataset \cite{shotton2013scene}.} 
{\rm Unsupervised~(Unsup.) methods cannot provide their scale and alignment; we applied them to the Sim(3) alignments estimated on the training dataset.
Pose Sup. represents the pose supervision (pGT: pseudo-ground-truth poses, \cite{orbslam3}: ORB-SLAM3, \cite{schonberger2016structure}: COLMAP.) is used.}
}
\label{tab:test}
\resizebox{1.0\hsize}{!}{
\begin{tabular}{c|l|c|ccccccc}
\hline
& Method & Pose & \multicolumn{7}{c}{Scene} \\
&        & Sup. & Chess & Fire & Heads & Office & Pumpkin & Redkitchen & Stairs \\
\hline
\multirow{8}{*}{\rotatebox[origin=c]{90}{\bf Supervised}}
& PoseNet2~\cite{kendall2017geometric} & pGT & 0.13m, 4.5$^{\circ}$	& 0.27m, 11.3$^{\circ}$	& 0.17m, 13.0$^{\circ}$	& 0.19m, 5.6$^{\circ}$		& 0.26m, 4.8$^{\circ}$		& 0.23m, 5.4$^{\circ}$	& 0.35m, 12.4$^{\circ}$		\\
& MapNet~\cite{brahmbhatt2018geometry} & pGT & 0.08m, 3.3$^{\circ}$	& 0.27m, 11.7$^{\circ}$	& 0.18m, 13.3$^{\circ}$	& 0.17m, 5.2$^{\circ}$		& 0.22m, 4.0$^{\circ}$		& 0.23m, 4.9$^{\circ}$			& 0.30m, 12.1$^{\circ}$		\\
& NN-Net~\cite{laskar2017camera} & pGT & 0.13m, 6.5$^{\circ}$	& 0.26m, 12.7$^{\circ}$	& 0.14m, 12.3$^{\circ}$	& 0.21m, 7.4$^{\circ}$		& 0.24m, 6.4$^{\circ}$		& 0.24m, 8.0$^{\circ}$			& 0.27m 11.82$^{\circ}$		\\
& ReLocNet~\cite{relocnet} & pGT & 0.12m, 4.1$^{\circ}$	& 0.26m, 10.4$^{\circ}$	& 0.14m, 10.5$^{\circ}$	& 0.18m, 5.3$^{\circ}$		& 0.26m, 4.2$^{\circ}$		& 0.23m, 5.1$^{\circ}$			& 0.28m, 7.5$^{\circ}$		\\
& CamNet~\cite{camnet} & pGT & 0.04m, 1.7$^{\circ}$	& 0.03m, 1.7$^{\circ}$	& 0.05m, 2.0$^{\circ}$	& 0.04m, 1.6$^{\circ}$		& 0.04m, 1.6$^{\circ}$		& 0.04m, 1.6$^{\circ}$			& 0.04m, 1.5$^{\circ}$		\\
& DSAC*~\cite{dsacstar} & pGT & 0.02m, 1.1$^{\circ}$	& 0.02m, 1.2$^{\circ}$	& 0.01m, $1.8^{\circ}$	& 0.03m, 1.2$^{\circ}$		& 0.04m, 1.4$^{\circ}$		& 0.03m, 1.7$^{\circ}$			& 0.04m, 1.4$^{\circ}$		\\
& \ - ORB-SLAM3  & \cite{orbslam3} & 1.23m, 38.6$^{\circ}$ & 1.28m, $59.1^{\circ}$	& 0.42m, 21.1$^{\circ}$	& 1.42m, 48.4$^{\circ}$	& 1.29m, 36.6$^{\circ}$	& 1.72m, 37.7$^{\circ}$	& 0.63m, 157.9$^{\circ}$	\\
& \ - COLMAP~\cite{brachmann2021limits} & \cite{schonberger2016structure} & 0.04m, 1.4$^{\circ}$ & 0.02m, 1.1$^{\circ}$ & 0.01m, 1.5$^{\circ}$ & 0.08m, 3.0$^{\circ}$ & 0.07m, 3.1$^{\circ}$ & 0.06m, 3.2$^{\circ}$ & 0.05m, 1.9$^{\circ}$ \\
\hline
\multirow{5}{*}{\rotatebox[origin=c]{90}{\bf Unsup.}} 
& ORB-SLAM3~\cite{orbslam3}           & & 0.91m, 27.3$^{\circ}$  & 0.92m, 29.3$^{\circ}$  & 0.17m, {\bf 5.0}$^{\circ}$  & {\bf 0.19}m, {\bf 5.5}$^{\circ}$  & 0.43m, 10.0$^{\circ}$  & 0.87m, 25.4$^{\circ}$  & 0.63m, 175.3$^{\circ}$     \\
& monodepth2{\dag}~\cite{monodepth2} & & 0.34m, 15.7$^\circ$ & 0.40m, 20.6$^\circ$ & 0.26m, 12.0$^\circ$ & 0.70m, 50.9$^\circ$ & 0.53m, 12.5$^\circ$ & 0.66m, 16.1$^\circ$ & 0.44m, 37.4$^\circ$ \\
& {\bf Ours (full)} & & {\bf 0.08}m, {\bf 3.1}$^{\circ}$ & {\bf 0.18}m, {\bf 7.2}$^{\circ}$	& {\bf 0.12}m, 7.8$^{\circ}$ & 0.56m, 39.6$^{\circ}$ & {\bf 0.18}m, {\bf 4.6}$^{\circ}$	& {\bf 0.21}m, {\bf 7.0}$^{\circ}$	& 0.42m, {\bf 27.5}$^{\circ}$	\\
& \ - w/o DSC & & 0.24m, 9.4$^\circ$ & 0.55m, 36.3$^\circ$ & 0.20m, 9.5$^\circ$ & 0.29m, 9.8$^\circ$ & 0.38m, 10.4$^\circ$ & 0.53m, 12.5$^\circ$ & {\bf 0.41}m, 38.6$^\circ$ \\
& \ - w/o LCVS & & 0.14m, 5.9$^\circ$ & 0.30m, 15.1$^\circ$ & 0.15m, 9.2$^\circ$ & 0.56m, 32.5$^\circ$ & 0.28m, 8.0$^\circ$ & 0.37m, 14.7$^\circ$ & 0.43m, 31.5$^\circ$ \\
\hline
\end{tabular}
}
\end{table*}
\begin{figure*}
\begin{center}
\includegraphics[width=1.0\hsize]{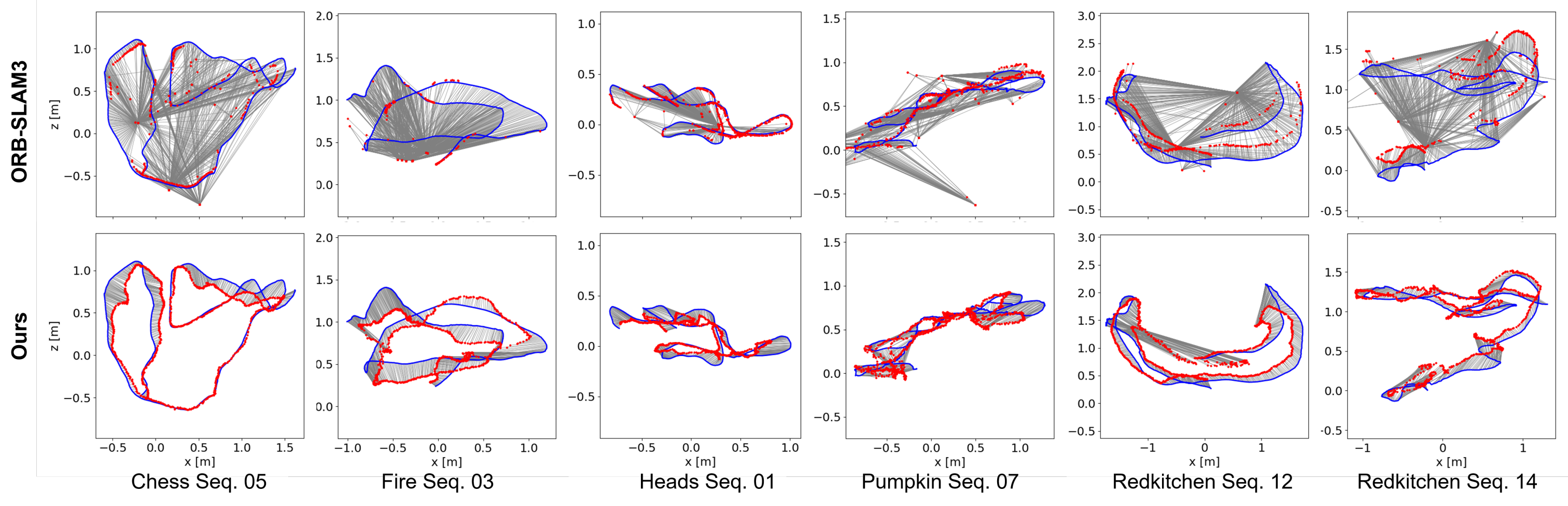}
\caption{{\bf Qualitative results of camera re-localization on 7-scenes test dataset \cite{shotton2013scene}.}
Blue lines represent pseudo-ground-truth camera trajectories, and red points represent predicted camera poses by our method and re-localization of ORB-SLAM3 \cite{orbslam3}.
Gray lines represent errors between predicted poses and corresponding pseudo-ground-truth poses.
We output the pose of the most matched key-frame in the candidates retrieved by BOW as an estimated pose when the camera re-localization of ORB-SLAM3 fails.
Therefore, many frames are matched to a key-frame pose which has much features.}
\label{fig:test}
\end{center}
\end{figure*}

\subsubsection{Camera Re-Localization Performance}
\label{sec:reloc}

First, we evaluate the camera re-localization performance using a 7-scenes test dataset \cite{shotton2013scene}.
At the test time, the final pose of our method is calculated by a median operator instead of the averaging in \eqref{eq:camera_pose} to achieve robustness against outliers.
For the comparison, we employ following baselines:

\vspace{1mm}
\noindent
{\bf ORB-SLAM3 \cite{orbslam3}}
ORB-SLAM3 equips the camera re-localization system to detect the camera pose when tracking with VO is lost.
The camera re-localization consists of image retrieval based on the bag-of-words (BOW) feature \cite{galvez2012bags} and pose optimization via RANSAC using ORB features \cite{rublee2011orb}.
In this evaluation, we executed ORB-SLAM3 on the training dataset of each scene to create a map and evaluate the re-localization performance on the test dataset.
The camera re-localization was not successful in all scenes; therefore, we output the pose of the most matched key-frame in the candidates retrieved by BOW as an estimated pose when the camera re-localization failed.

\vspace{1mm}
\noindent
{\bf DSAC* \cite{dsacstar} (pose sup. \cite{orbslam3}, \cite{schonberger2016structure})}
We trained DSAC* with pose supervision via ORB-SLAM3 \cite{orbslam3}, and listed the results of DSAC* trained with pose supervision via COLMAP \cite{schonberger2016structure} in \cite{brachmann2021limits} as a reference method of the supervised camera re-localization with SLAM and SfM from RGB sequences.

\vspace{1mm}
\noindent
{\bf monodepth2{\dag} \cite{monodepth2} }
We trained monodepth2 for unsupervised camera re-localization to estimate the camera pose from a single image instead of a relative camera pose in the original monodepth2.
Following our method, we calculated the relative pose from estimated camera poses to predict images. 
This is also an ablation study to confirm the effectiveness of our unsupervised learning framework with the directed scene coordinates and loop closed view synthesis.

\vspace{2mm}
Unsupervised camera re-localization methods do not provide scale and alignment of trajectories; therefore, we applied Sim(3) alignments estimated on the training dataset using {\it evo} \cite{grupp2017evo} for testing.
The median re-localization error of position (m) and attitude ($^\circ$) on the 7-scenes test dataset are shown in Table \ref{tab:test}, which also lists the median errors of conventional supervised camera re-localization methods. 

Based on the results of unsupervised camera re-localization, our method outperforms all competitors in most scenes.
ORB-SLAM3 camera re-localization cannot provide accurate results because the camera re-localization fails in most frames of each scene.

The monodepth2{\dag} is less effective than our method for all scenes.
This result confirms the contribution of our learning framework for camera re-localization based on the directed scene coordinates and loop closed view synthesis.

Performance of our method degrades on the `Office' and `Stairs'.
This is because there is a similar view on different poses in the scenes.
`Office' has similar desks put in different positions in the room.
`Stairs' are very repetitive sequences because the camera goes up or down the stairs.
Addressing these repetitive structures will be the scope of our future work.

The qualitative results of camera re-localization are shown in Fig. \ref{fig:test}.
As can be seen from the figures, our method can predict camera poses consistently in all frames.
In contrast, ORB-SLAM3 cannot perform camera re-localization on many frames, the most matched candidates which are the alternative outputs cause the degradation in the performance.
These results suggest the robustness of our camera re-localization.

Comparing the supervised and unsupervised methods, we can see how challenging unsupervised camera re-localization is.
Nevertheless, our methods outperformed several previous supervised camera re-localization methods, such as PoseNet2 \cite{kendall2017geometric}, MapNet \cite{brahmbhatt2018geometry}, NN-Net \cite{laskar2017camera}, and ReLocNet \cite{relocnet}, in most scenes.
Although it does not achieve the performance of state-of-the-art supervised camera re-localization, such as CamNet \cite{camnet} and DSAC* \cite{dsacstar}, the fact that our unsupervised learning can achieve an accuracy approaching that of supervised camera re-localization is a significant achievement.

From this result, the DSAC*\cite{dsacstar} can provide accurate results with COLMAP~\cite{schonberger2016structure}, however, it cannot be trained correctly with ORB-SLAM3~\cite{orbslam3}.
This denotes the performances of supervised methods depend on the accuracy of localization methods.
Our method does not achieve the accuracy of DSAC* based on COLMAP, however, it facilitates estimating the depth accurately online.

\begin{table*}
\caption{
{\bf Quantitative results for depth-estimation performance.}
{\rm Comparison of our method to monodepth2 \cite{monodepth2} on 7-scenes test dataset \cite{shotton2013scene}.
The predicted depth maps are scaled by a scalar that matches the median with the ground-truth for each frame.
Standard deviations per median (std/med) of scale factors representing scale consistency are also shown.
The evaluation was performed in the range of $0.1$ m - $10$ m, and averaged over scenes.}}
\label{tab:depth}
\resizebox{1.0\hsize}{!}{
\begin{tabular}{l|c|cccc|ccc}
\hline
Method & Scale factor 	& \multicolumn{4}{|c|}{Error metric} & \multicolumn{3}{|c}{Accuracy metric}						\\
& std / med		& Abs Rel & Sq Rel & RMSE & RMSE log & $\delta < 1.25$ & $\delta < 1.25^2$ & $\delta < 1.25^3$ 	\\
\hline
monodepth2\cite{monodepth2} & 0.080 & 0.212 & 0.412 & 0.638 & 0.275 & 0.761 & 0.908 & 0.958 \\
monodepth2{\dag} & 0.044 & 0.220 & 0.445 & 0.663 & 0.270 & 0.748 & 0.913 & 0.961 \\
\bf{Ours (full)} & {\bf 0.017} & {\bf 0.135}  & {\bf 0.059} & {\bf 0.308} & {\bf 0.178} & {\bf 0.825} & {\bf 0.962} & {\bf 0.992} \\
\ - w/o DSC & 0.030 & 0.171 & 0.164 & 0.432 & 0.218 & 0.771 & 0.942 & 0.982 \\
\ - w/o LCVS & 0.022 & 0.145 & 0.065 & 0.325 & 0.189 & 0.798 & 0.959 & 0.991 \\
\hline
\end{tabular}
}
\end{table*}

We also listed results of our ablation study: 

\vspace{1mm}
\noindent
{\bf - w/o DSC}
This represents our method without a directed scene coordinate network~(DSC).
The pose network is the same as in monodepth2{\dag}.

\vspace{1mm}
\noindent
{\bf - w/o LCVS}
This represents our method without loop closed view synthesis~(LCVS).
The photometric re-projection losses are calculated from two relative poses between three images as same as monodepth2{\dag}\cite{monodepth2}.

\vspace{2mm}
The benefits of DSC and LCVS are clearly shown in Table \ref{tab:test}, respectively.
Focusing on each scene, we can see that the effectiveness of both methods is different for each scene.
The effectiveness of DSC on `Fire' and `Pumpkin' is better than that of LCVS, however, DSC suffers from repetitive environments such as `Office' and `Stairs.'
As mentioned above, addressing these repetitive structures will be the scope of future work.
Although the best method is different for each scene, our full method achieved the best average performance by combining DSC and LCVS.

\subsubsection{Depth Estimation Performance}

Next, we evaluated the depth estimation performance using ground-truth depth in the 7-scenes dataset \cite{shotton2013scene}.
In this evaluation, we compared our method to a recent monocular depth estimation method, monodepth2 \cite{monodepth2}, trained on the 7-scenes dataset as follows:

\vspace{1mm}
\noindent
{\bf monodepth2 \cite{monodepth2}}
Monodepth2 evaluated here is trained by all train sequences for 7-scenes, because it is difficult to train by sequences from each scene respectively.
The training was performed in 500 epochs. 
The monodepth2 uses three-frame image sequences for training; however, the sampling rate of the 7-scenes dataset is too high to train monodepth2.
Therefore we generate a three-frame image sequence within $10$ step intervals for training.

\vspace{2mm}
Table \ref{tab:depth} shows the evaluation results of the depth estimation using the 7-scenes test dataset.
Because neither model provides the depth scale, we multiply the predicted depth maps by a scalar that matches the median with the ground-truth for each frame.
The relative standard deviations of the scale factors are listed in Table \ref{tab:depth}.

As shown in Table \ref{tab:depth}, our models outperformed monodepth2 on all metrics.
It should be noted that with regards to the standard deviation of the scale factor, our model has a much smaller relative standard deviation of scale factors than the monodepth2 model, indicating that our models can predict depth with a more consistent scale across frames.
Thus, our method can predict a better and more consistent depth than monodepth2 in known scenes.

We also performed an ablation study for the depth estimation.
The benefits of DSC and LCVS for the depth estimation are also clearly shown in Table \ref{tab:depth}, respectively.
Our full method achieved the best performance by combining DSC and LCVS.

\begin{figure}
\begin{center}
\includegraphics[width=1.0\hsize]{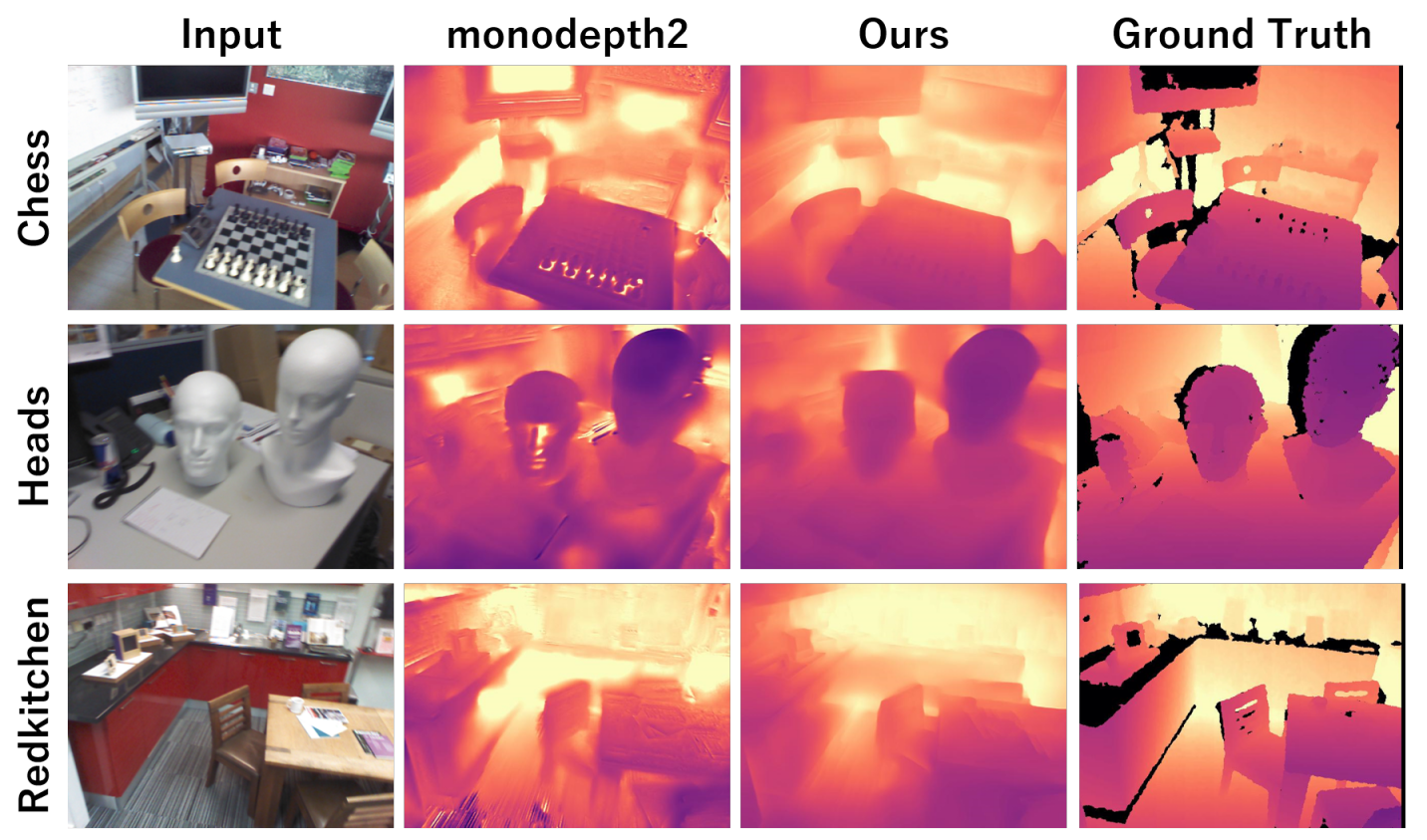}
\caption{{\bf Qualitative results of depth on 7-scenes test dataset \cite{shotton2013scene}.}
Our method has fewer artifacts than monodepth2 \cite{monodepth2}.
}
\label{fig:depth}
\end{center}
\end{figure}

The qualitative results are shown in Fig. \ref{fig:depth}, which shows our method has fewer artifacts than monodepth2 \cite{monodepth2}.
These results suggest that the proposed framework can train a depth network with better performance than a recent monocular depth estimation for indoor scenes.
Note that the depth estimation of the proposed framework is trained jointly with camera re-localization; therefore, the depth model is also limited to the trained scene.

\section{Conclusion}
In this study, we present an unsupervised simultaneous learning framework for camera re-localization and depth estimation from unlabeled video sequences.
Our method simultaneously trained two networks for directed scene coordinates and depth map by view synthesis via a video sequence.
Camera re-localization is performed by geometrically combining the directed scene coordinates and the depth map.
To the best of our knowledge, this is the first study to propose a camera re-localization training method without supervised learning.

We evaluated our method on the 7-scenes dataset \cite{shotton2013scene}, and demonstrated its performance in camera re-localization and depth estimation.
The results show that our method outperformed the camera re-localization embedded in ORB-SLAM3 \cite{orbslam3}.
Moreover, despite unsupervised learning, our method outperforms several previous supervised camera re-localization methods.
In addition, it also outperformed monodepth2 \cite{monodepth2} on monocular depth estimation in known environments.

However, the current learning framework is limited to frequent and small indoor camera sequences, because the learning framework depends on the overlap of the viewed scene between three or more camera images.
Thus, future research will aim at improving the method for applicability to wider scenes and data.

%\clearpage

\bibliographystyle{IEEEtran}
\vskip-\parskip
\begingroup
\footnotesize
\bibliography{root}
\endgroup

\end{document}